\documentclass{article}
\usepackage[nonatbib,preprint]{neurips_2022}

\usepackage{graphicx}
\usepackage{amsmath}
\usepackage{amssymb}
\usepackage{booktabs}
\usepackage{tabularx}
\usepackage{xcolor}
\usepackage{soul}

\DeclareMathOperator*{\argmin}{arg\,min}
\newcommand{\our}{{MAPLE}}
\newcommand{\Lagr}{\mathcal{L}}

\usepackage[breaklinks,colorlinks]{hyperref}

\usepackage[
backend=biber,
sorting=ynt
]{biblatex}

\usepackage[capitalize]{cleveref}
\crefname{section}{Sec.}{Secs.}
\Crefname{section}{Section}{Sections}
\Crefname{table}{Table}{Tables}
\crefname{table}{Tab.}{Tabs.}

\title{MAPLE: Microprocessor A Priori for Latency Estimation}
\author{Saad Abbasi\\
Waterloo AI Institute\\
University of Waterloo\\
Waterloo, Ontario, Canada\\
\and
Alexander Wong\\
Waterloo AI Institute\\
University of Waterloo\\
DarwinAI\\
Waterloo, Ontario, Canada\\
\and
Mohammad Javad Shafiee\\
Waterloo AI Institute\\
University of Waterloo\\
DarwinAI\\
Waterloo, Ontario, Canada\\
}
\begin{document}

\maketitle

\begin{abstract}
   Modern deep neural networks must demonstrate state-of-the-art accuracy while exhibiting low latency and energy consumption. As such, neural architecture search (NAS) algorithms take these two constraints into account when generating a new architecture. However, efficiency metrics such as latency are typically hardware dependent requiring the NAS algorithm to either measure or predict the architecture latency. Measuring the latency of every evaluated architecture adds a significant amount of time to the NAS process. Here we propose Microprocessor A Priori for Latency Estimation (MAPLE) that leverages hardware characteristics to predict deep neural network latency on previously unseen hardware devices. MAPLE takes advantage of a novel quantitative strategy to characterize the underlying microprocessor by measuring relevant hardware performance metrics, yielding a fine-grained and expressive hardware descriptor. The CPU-specific performance metrics are also able to characterize GPUs, resulting in a versatile descriptor that does not rely on the availability of hardware counters on GPUs or other deep learning accelerators. We provide experimental insight into this novel strategy. Through this hardware descriptor, MAPLE can generalize to new hardware via a few shot adaptation strategy, requiring as few as 3 samples from the target hardware to yield 6\% improvement over state-of-the-art methods requiring as much as 10 samples. Experimental results showed that, increasing the few shot adaptation samples to 10 improves the accuracy significantly over the state-of-the-art methods by 12\%. We also demonstrate MAPLE identification of Pareto-optimal DNN architectures exhibit superlative accuracy and efficiency. The proposed technique provides a versatile and practical latency prediction methodology for DNN run-time inference on multiple hardware devices while not imposing any significant overhead for sample collection.
\end{abstract}

\section{Introduction}
\label{sec:intro}
In the previous decade, deep neural networks (DNNs) have been widely used with great efficacy for a variety of tasks including computer vision~\cite{krizhevsky2012imagenet,simonyan2014very,szegedy2015going,he2016deep}, natural language processing~\cite{sutskever2014sequence,vaswani2017attention}, and speech recognition~\cite{graves2013speech}. However, designing state-of-the-art DNNs is a time-consuming process, often requiring iterative training and validation to ensure the model meets the target accuracy requirements. Furthermore, applications which require on-device inference (e.g. privacy-preserving facial recognition, autonomous driving) exacerbate this process as the task-specific DNNs must now satisfy multiple constraints of energy consumption, inference latency or memory footprint, in addition to just the accuracy. In recent years, algorithmic solutions such as neural architecture search (NAS)~\cite{zoph2016neural,baker2016designing,pham2018efficient,liu2018darts,liu2018progressive,xu2019pc,chen2020drnas} have received significant attention to automatically find Pareto-optimal architectures that achieve superlative efficacy and accuracy simultaneously~\cite{tan2019mnasnet, berman_aows_2020, cai2018proxylessnas, wu2019fbnet, wan2020fbnetv2, dai2019chamnet, xu2020latency, cai2019once, zhang2020fast}. The model efficiency is typically measured via hardware dependent metrics such as architecture latency, memory or energy consumption, which are computationally expensive to acquire.

 Nonetheless, achieving low latency is paramount for applications that require real-time feedback or need to preserve privacy. Since DNN latency is dependent on the network architecture as well as the underlying hardware, discovering Pareto-optimal architectures becomes highly challenging due to the diverse number of hardware devices, accelerators and frameworks available for servers and edge devices. To mitigate this issue, some NAS approaches execute the evaluated architectures on the target device to measure the true architecture latency~\cite{tan2019mnasnet}. However, this approach quickly becomes challenging to scale due to i) the large number of architectures that need to be evaluated (e.g. ProxylessNAS evaluates 300,000 architectures in its first round) and ii) the diversity of available hardware (e.g. CPUs, GPUs, ASICs). As an example, if we assume the average latency of a DNN architecture is 50 ms and we take 50 measurements to reduce variance, Proxyless NAS would need 750 hours, or nearly a month of continuous data collection, to evaluate 300,000 architectures. This is clearly prohibitive for even earlier NAS approaches which required immense computational resources. More recent NAS approaches, such as DARTS, are able to discover optimal architectures in a few hours \cite{liu2018darts}. The efficiency of recent NAS approaches necessitates a need for a practical hardware-aware NAS that can mitigate the large sample requirement and enables rapid adaptation to different hardware devices. 
 
\begin{figure*}
  \centering
    \includegraphics[width=1.0\textwidth]{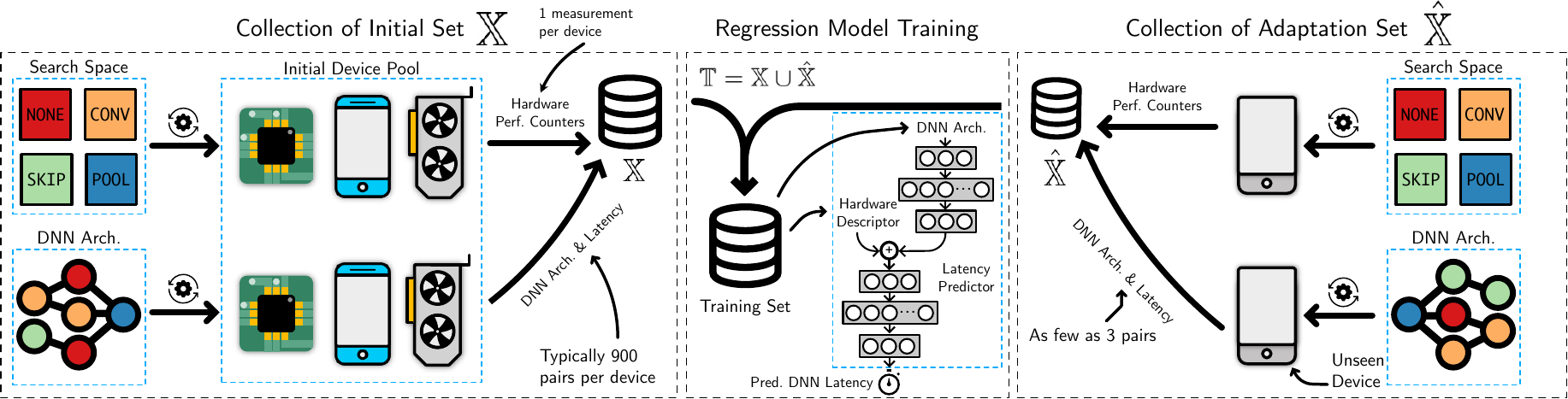}
    \caption{{\our} overview; {\our} is a hardware-aware latency predictor capable of inferring DNN architecture latency on new (previously unseen) devices. {\our} relies on a novel quantitative characterization of the device microprocessor through hardware performance counters. These quantitative metrics characterize the target device based on CPU cache efficacy, computational speed, and memory I/O among other events. {\our} collects these metrics while executing the NAS search space operations yielding a fine-grained descriptor capable of discerning between different hardware. {\our} also measures the latency of an initial set of DNN architectures on a set of known devices, enabling generalization on unseen DNN architectures. To adapt to previously unseen hardware, {\our} characterizes the new hardware through the fine-grained performance metric-based hardware descriptor and measures the latency of just 3 randomly chosen architectures. The few shot adaptation sample efficacy is primarily due to the effectiveness of the quantitative strategy to characterize hardware. The training set is formed by mixing the three adaptation samples into the initial set. The latency predictor is implemented as a neural network-based regression model which is fed DNN architecture encodings and hardware descriptors.}
     \label{fig:overview}
\end{figure*}

Motivated by these challenges, we propose Microprocessor A Priori for Latency Estimation ({\our}). As shown in  Figure~\ref{fig:overview}, {\our} is a hardware-aware latency predictor based on a novel processor prior modeling strategy for quantitatively characterizing the underlying processor. This hardware descriptor characterizes the target devices' main processor through vendor-provided hardware performance counters. Performance counters monitor events related to process execution, such as the number of instructions, cycles, branch predictions or cache hits and misses, among hundreds of other similar events. An important consideration here is the availability of performance counters on different hardware. CPU vendors have included performance monitoring units on a wide variety of devices for over a decade~\cite{treibig2010likwid,liu2010survey,mathur2003toward}. CPU-based performance monitoring is widely used to analyze software performance. In contrast, GPU vendors only support performance monitoring units on a few devices. To avoid being limited by performance-counter availability, MAPLE only relies on microprocessor performance monitoring units. More specifically, MAPLE uses CPU performance monitoring events to characterize GPU hardware by taking advantage of the tight I/O coupling between the CPU and GPU, particularly in latency-oriented applications. This yields a versatile technique that can characterize a wide variety of hardware.

MAPLE measures performance counters while executing all possible operations in the NAS search space. We postulate that characterizing the system hardware at the operator level (rather than at the architecture level) leads to two main benefits. First, even though a NAS search space can create a large number of distinct architectures, the search space is usually comprised of only a few fixed, stable primitive operators with only a few possible parameter choices. The small number of primitive operators results in a relatively small but expressive hardware descriptor, independent of the possible number of architectures in the target search space. Second, since all architectures are comprised of operations from the search space, operator-level characterization would be expressive enough to generalize more effectively to unseen architectures and hardware, yielding higher accuracy while requiring less samples.

Through such operator-level characterization of the hardware, we demonstrate that the proposed {\our} algorithm adapts to new computing hardware with as few as 3 samples collected from the device. The effectiveness of the hardware descriptor also precludes the need for model adaptation techniques such as meta-learning or transfer-learning. Instead, {\our} simply relies on mixing the measurements from the new hardware during training.

We demonstrate the adaptive strength of {\our} on eight different devices (3 CPUs and 5 GPUs). To summarize, the key contributions of this study are as follows:
\begin{itemize}
  \item We introduce a hardware-aware latency predictor based on a novel microprocessor prior modeling strategy for quantitatively characterizing the underlying processor through hardware performance monitoring events.
   \item By taking advantage of the tight-coupling between a CPU and GPU, we characterize GPU behavior using CPU-based hardware performance counters.
  \item We propose a simple latency prediction technique that is able to generalize to new hardware with only 3 measurements.
  \item By characterizing the hardware effectively via the novel approach, we are able to infer latency with higher accuracy while requiring a less diverse hardware pool compared to the state-of-the-art algorithms.
\end{itemize}

\section{Related Works}

Several research ideas have aimed to alleviate the time and engineering effort required to predict DNN architecture latency on a target computing device. One of the simplest ways to exploit hardware characteristics and use hardware-aware NAS is by employing FLOPs as a proxy for on-device latency estimation~\cite{he2018amc,li2016pruning}. However, FLOPs are in general, too simple of a measure as most deep learning operations are not compute-bound. FLOPs typically lead to inaccurate estimation of on-device latency. Another simple technique of estimating the latency is by building an on-device latency look-up table (LUT) of all possible operations in a given search space~\cite{dai2019chamnet,cai2019once,wu2019fbnet,cai2018proxylessnas}. The NAS then sums the relevant operator latency to estimate the execution time of a given architecture. However, a latency LUT fails to capture the intricacies of architecture optimization (such as layer fusion or I/O optimization) and thus typically exhibits a deviation of 10-20\% from the actual end-to-end latency, depending on the underlying hardware and architecture. 

A promising approach is to train a regression model on a dataset of architecture and latency pairs, collected from the target device~\cite{wang2020hat,dudziak2020brp,liu2021fox}. The regression model can predict the latency of unseen architectures and removes the need to measure the latency of every architecture during the NAS process; significantly decreases the amount of time spent on acquiring latency measurements. This approach results in significantly lower error compared to LUT-based or FLOP-based approaches. However, the regression-model approaches quickly becomes difficult to scale since it requires the NAS algorithm to train a new model for every target hardware, necessitating the collection of a large number of architectures and latency pairs from all target devices.

 Building upon latency predictors, Syed and Srinivasan~\cite{syed2021generalized} used transfer learning to adapt a regression model to unseen hardware. Although transfer learning reduces the overall cost of sample collection from new hardware, it still requires a considerable number of measurements (approx. 700 samples) from unseen hardware. Towards predicting latency on new hardware, Lee \textit{et al.}~\cite{lee2021help} employed meta-learning techniques to develop HELP. Similar to our work, HELP adapts a regression model to infer latency on previously unseen hardware. The rapid adaptation is mainly due to the characterization of the hardware using end-to-end latency of reference architectures. Although this technique demonstrates state-of-the-art performance, it requires at least 10 measurements from new hardware to adapt effectively and does not explicitly characterize the hardware.

\section{Methodology}
Our goal is to design a latency estimation inference model capable of generalizing to different hardware by a few shot adaption strategy. To this end, in this section we formulate the hardware-aware regression model, explain the hardware descriptor and provides details of the dataset collection pipeline.

\subsection{Problem Formulation}
\label{sec:problem}
Although DNN Latency is a function of network architecture as well as the underlying hardware architecture, most latency oriented NAS approaches~\cite{syed2021generalized,dudziak2020brp} model DNN latency solely as a function of network architecture. Such techniques are incapable of generalizing across hardware devices and therefore require building a predictor for each target device. Moreover, to generalize across different DNN architectures, we need to collect a large number of architecture and latency pairs from each target device.

 To enable rapid adaptation to new hardware devices, the latency model must take into account some hardware characteristics (i.e. be hardware-aware) as prior knowledge. To this end, we formulate the problem of inference latency estimation as a hardware-aware regression model. Formally, the hardware-aware regression model can be defined as $f\left(\textbf{a},\textbf{S};\theta\right)\mapsto \hat{y}$ where $\textbf{a}$ is the architecture encoding vector, $\textbf{S}$ is the quantitative hardware descriptor and $\hat{y}$ is the predicted latency. This formulation enables the regression model to predict $\hat{y}$ on different hardware ($\textbf{S}$) for the same architecture $\textbf{a}$.

Most hardware-agnostic latency estimation techniques train the regression model on architecture encodings and subsequently adapt it to different hardware through some form of domain adaptation~\cite{lee2021help,syed2021generalized}. In contrast, {\our} adapts to new hardware at training time. The training set consists of latency measurements from 900 architectures collected from seven initial devices. This training set is augmented by a few latency measurements (as few as 3) from the target device. Formally,  we define the initial set of samples ($\mathbb{X}$) as
\begin{align}
    \mathbb{X} = \Big\{ \big(\textbf{a},\textbf{S},y\big)\mid \textbf{a}\in \mathbb{A}, \textbf{S}\in \mathbb{S}, y\in \mathbb{Y}\Big\}
\end{align}

where $\textbf{S}$ is the set of hardware descriptors characterizing every hardware in the training device-pool, $\mathbb{A}$  is a set of architectures (typically 900) for which we collect the on-device latency, $\mathbb{Y}$, from each device in the training device-pool. Similarly, the adaptation set $\mathbb{\hat{X}}$ can be described as:
\begin{align}
    \mathbb{\hat{X}} = {\Big\{\big(\textbf{a},\textbf{S},y\big)\mid \textbf{a}\in \mathbb{\hat{A}}, \textbf{S}\in \mathbb{\hat{S}}, y\in \mathbb{\hat{Y}}}\Big\}
\end{align}
where $\mathbb{\hat{Y}}$ is the measured latency belonging to a set of randomly selected DNN architectures $\mathbb{\hat{A}}$, and $\mathbb{\hat{S}}$ is a set of hardware descriptors characterizing the previously unseen target hardware. The random adaptation architectures are sampled from the entire NAS-Bench-201 search space and are not restricted to the 900 training architectures used for the initial set. The actual training set is simply $\mathbb{T}=\mathbb{X} \cup \mathbb{\hat{X}}$. Since the number adaption examples is considerably less than the number of initial examples, we minimize weighted mean absolute error $\Lagr$ over the training set $\mathbb{T}$:
\begin{align}
    \argmin_{\theta} = \Lagr\Big(f\big(\textbf{a},\textbf{S};\theta\big),w,Y\Big)
\end{align}
where $w$ is the assigned sample weight. The adaptation samples are assigned a weight of $\frac{1}{\sqrt{\lvert\mathbb{\hat{X}}\rvert}}$ and the initial set of examples are assigned a weight of $\frac{1}{\sqrt{\lvert\mathbb{X}\rvert}}$. This weighting scheme ensures that the regression model prioritizes the samples from new hardware during training.

The architecture $\textbf{a}$ is encoded using a one-hot encoded operations matrix~\cite{white2020study} which defines each edge operation in a given architecture. The hardware descriptor $\textbf{S}$ is captured by measuring 10 distinct hardware performance counters while executing all possible operations defined in the search space. The hardware descriptor thus effectively carries a unique characterization between the search space and the underlying device. We discuss the hardware descriptor $\textbf{S}$ in detail in \cref{sec:descriptor}.

\subsection{Dataset and Latency Collection Pipeline}
\label{sec:pipeline}
Following experimental setup in BRP-NAS~\cite{dudziak2020brp} and HELP~\cite{lee2021help}, we use the NAS-Bench-201 dataset~\cite{dong2020bench} for our experiments. NAS-Bench-201 defines 15,625 cell-based DNN architectures and provides accuracy figures for CIFAR-10, CIFAR-100 and ImageNet-16-120 datasets. A NAS-Bench-201 cell is a 4-node densely connected directed acyclic graph. Within each cell, each edge is associated with an operation from the NAS-Bench-201 search space. The search space includes five possible operations including none, skip-connection, conv1x1, conv3x3 and avgpool3x3. Each operation can take on a channel width of 16, 32 and 64, yielding a total of 15 possible operation types.
 
We characterize a given hardware device by measuring key performance metrics while executing each operation in the NAS-Bench-201 search space (further details in Section~\ref{sec:descriptor}). More specifically, by employing the Linux performance analysis tool \emph{perf}~\cite{perf}, we measure key hardware performance metrics while executing each of the 15 operations in the search space. Given that we measure 10 hardware counters per descriptor, the resulting size of the hardware descriptor $\textbf{S}$ is 150. In addition to these performance metrics, we also measure the latency of each operation. The rationale for including the operator latency is to allow the model to correlate the end-to-end architecture latency with the operator latency.

In addition to hardware characterization, we also measure the end-to-end latency of all 15,525 architectures in NAS-Bench-201 on eight devices. These devices include an Intel Core i5-7200k, an Intel i9-9920k, an Intel Xeon Gold 6230, a Nvidia GTX-1070, a Nvidia RTX-2080 Ti, an Nvidia TitanX, an Nvidia TitanXP and finally an Nvidia RTX-6000. Each architecture latency is a mean of 50 measurement runs to reduce variance.

\subsection{Quantitative A Priori Hardware Descriptor}
\label{sec:descriptor}

\begin{figure}
    \includegraphics[width=\linewidth]{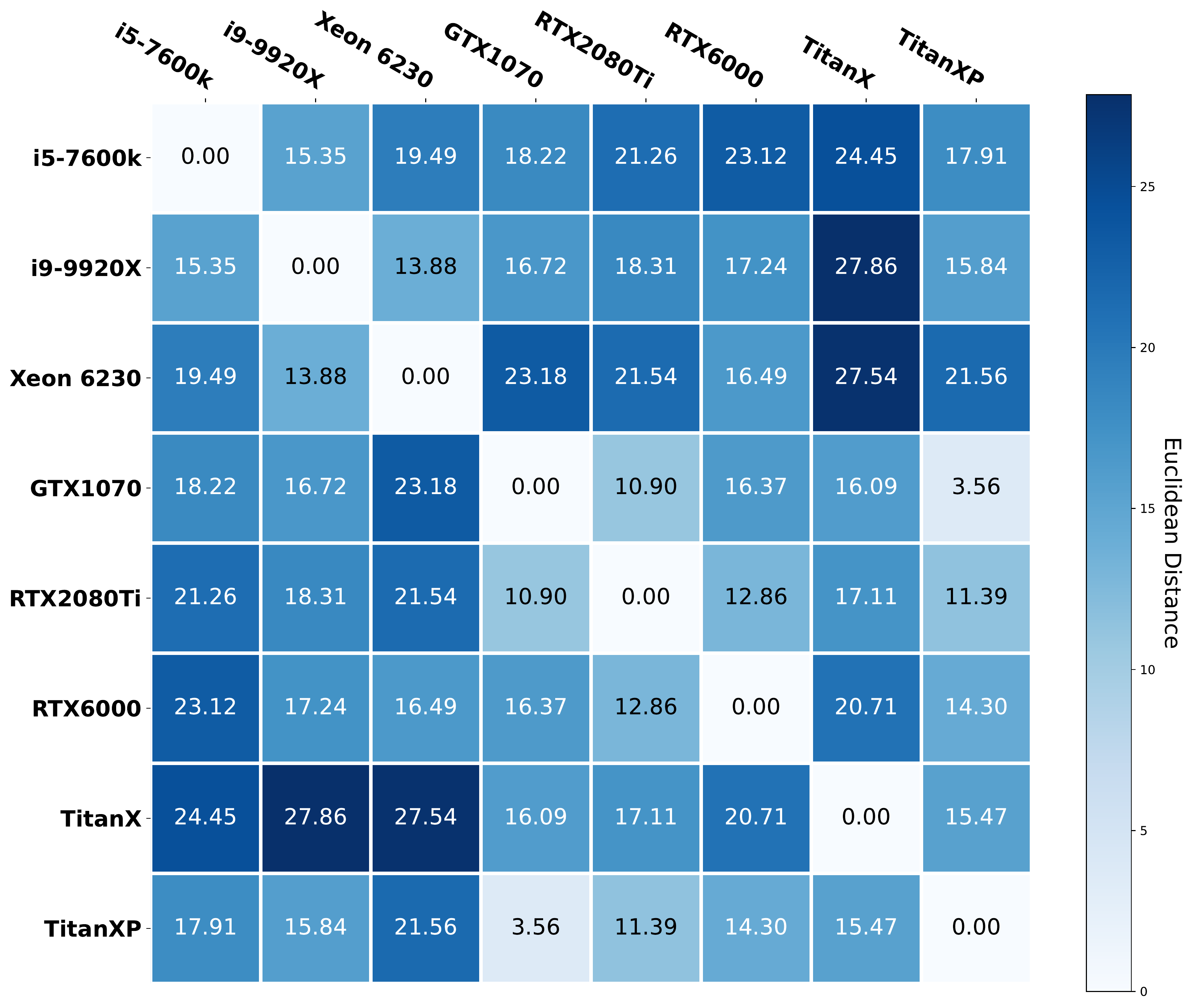}
\caption{A distance map illustrating the dissimilarity between hardware descriptors. Of particular note is the distance between the Titan X and Titan XP, both of which are paired with the same CPU. Despite using the same CPU (i.e. proxy device), CPU-specific hardware counters are different for each GPU due to the tight I/O coupling between the microprocessor and GPU.}
\label{fig:cosine}
\end{figure}

The quantitative a priori model described in Section~\ref{sec:problem} requires a hardware descriptor to characterize hardware devices. An effective hardware descriptor would uniquely parameterize different devices, enabling the regression model to distinguish between various hardware. We construct this hardware descriptor via various hardware performance counters which characterize workload execution. Hardware performance counters are special-purpose hardware registers within microprocessors that track events related to CPU-cycles, instruction counts, branch mispredictions, and cache miss rates, among other vital low-level metrics. These metrics are widely used for fine-grained performance analysis and for identifying bottlenecks within programs \cite{hpcmemory,hpcmemory2}.

We identify ten different hardware performance counters that can characterize the hardware effectively. These counters include CPU-cycles, instructions, cache-references, cache-misses, level one (L1) data cache loads, L1 data cache load misses, last-level cache (LLC) load misses, LLC loads, LLC store-misses and LLC stores. These event counters characterize if a given workload is compute or memory bound, how effective is the cache utilization and how often the system needs to request data from the main memory. We emphasize that the hardware descriptor is not used to predict the performance on a target device, it is merely used to parameterize the hardware such that it leads to a rich representation in the latent space. An important consideration here is the workload executed while the above-mentioned hardware counters are monitored. Some possible options include executing several architectures or a set of reference architectures that can represent the underlying architecture-latency distribution. However, identifying reference architectures becomes challenging as the number of possible architectures allowed by a NAS search space grows. Similarly, the number of architectures that may need to be executed would also increase with the number of possible architectures. In lieu of using DNN architectures as workloads, we propose a more fine-grained approach where we execute all operations in the NAS-Bench-201 search space. The advantage offered by this approach is that its independent of the number of architectures and only depends on the search space size.

\begin{figure}
\vspace{-1 cm}
\begin{tabularx}{\linewidth}{cc}
\setlength{\tabcolsep}{0.01cm} 
    \textbf{Intel Xeon 6230} & \textbf{Nvidia RTX 6000}\\ \hline
     \includegraphics[width=0.45\linewidth]{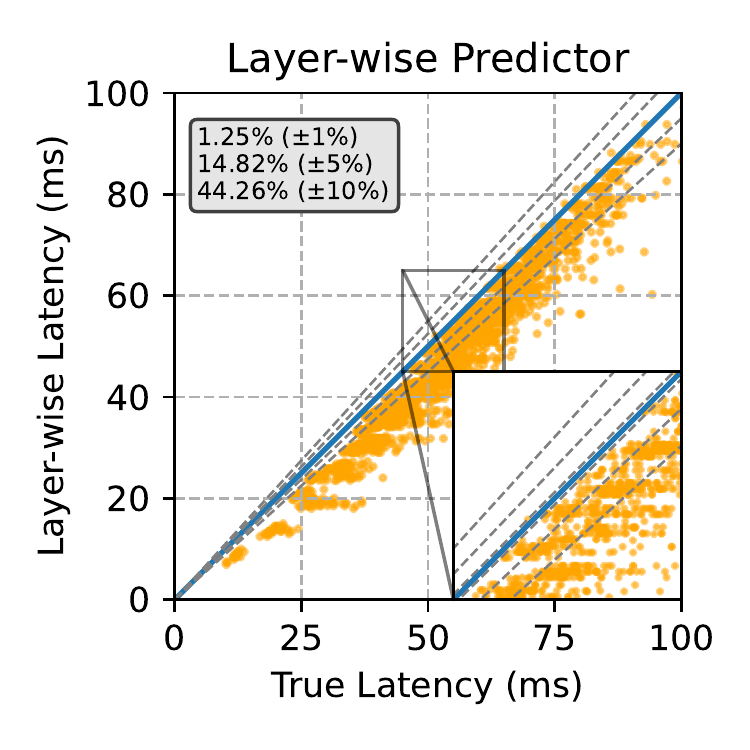} &  \includegraphics[width=0.45\linewidth]{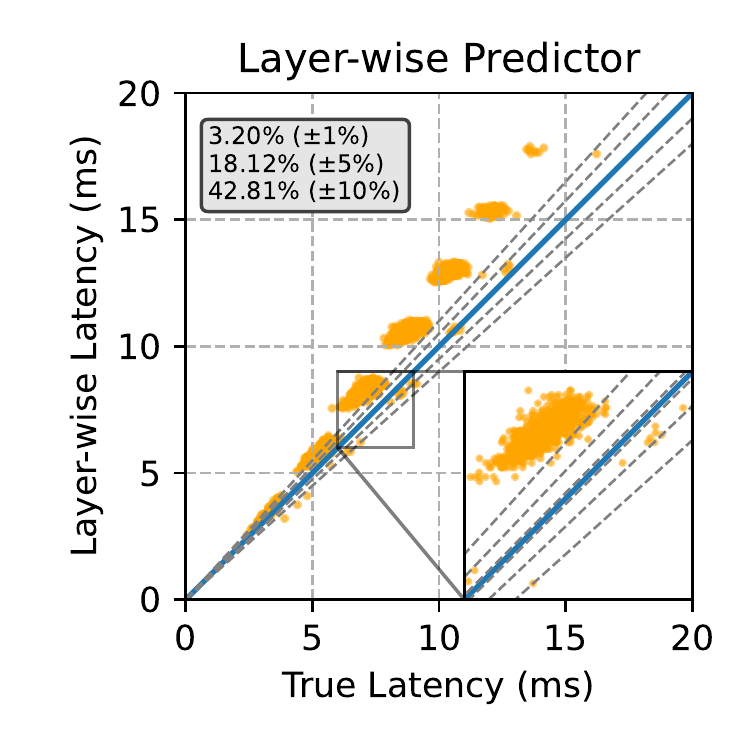}\\
     \includegraphics[width=0.45\linewidth]{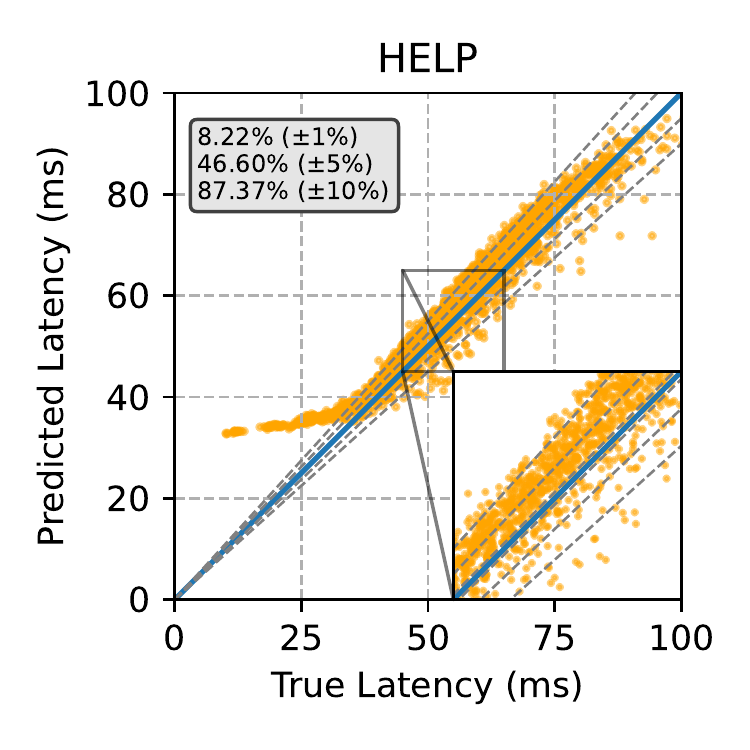} & \includegraphics[width=0.45\linewidth]{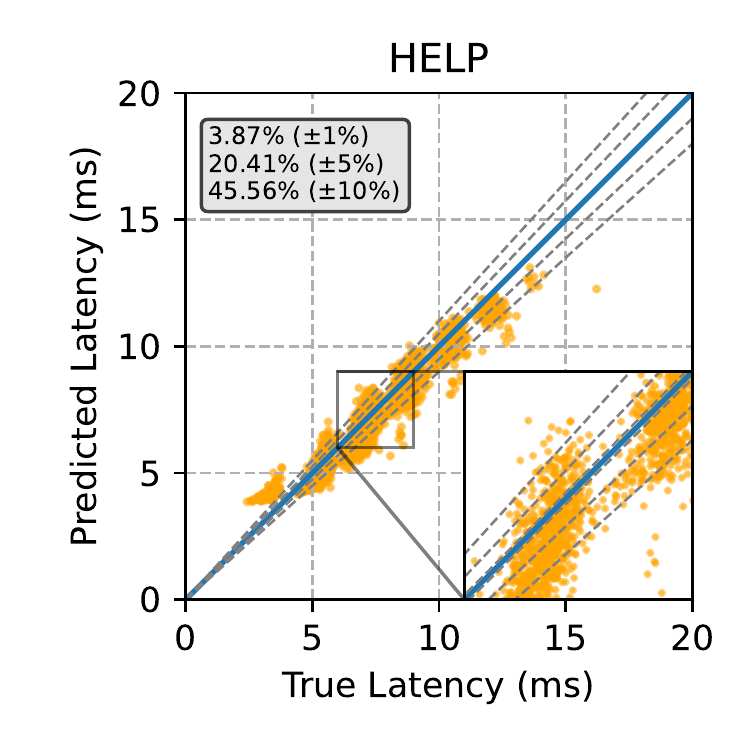} \\
     \includegraphics[width=0.45\linewidth]{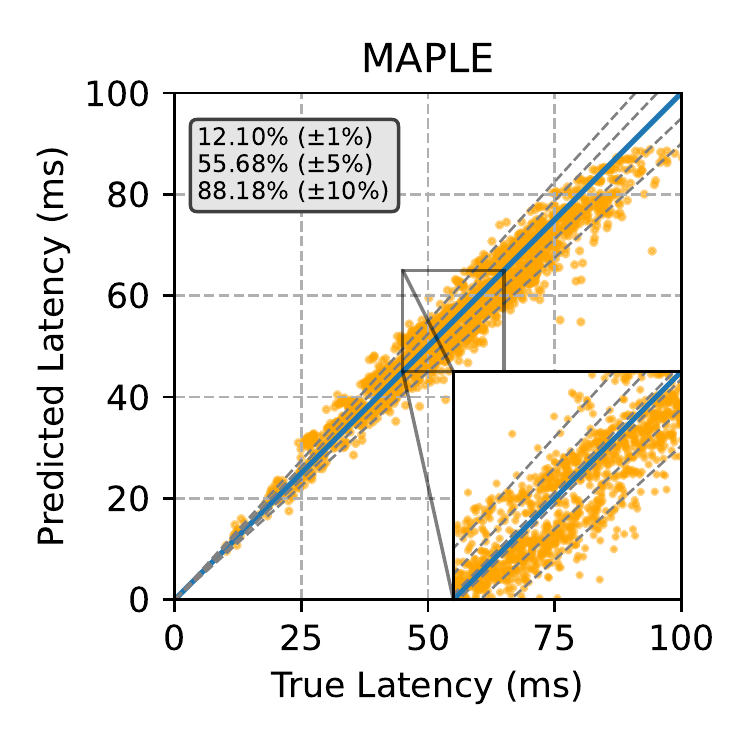} & \includegraphics[width=0.45\linewidth]{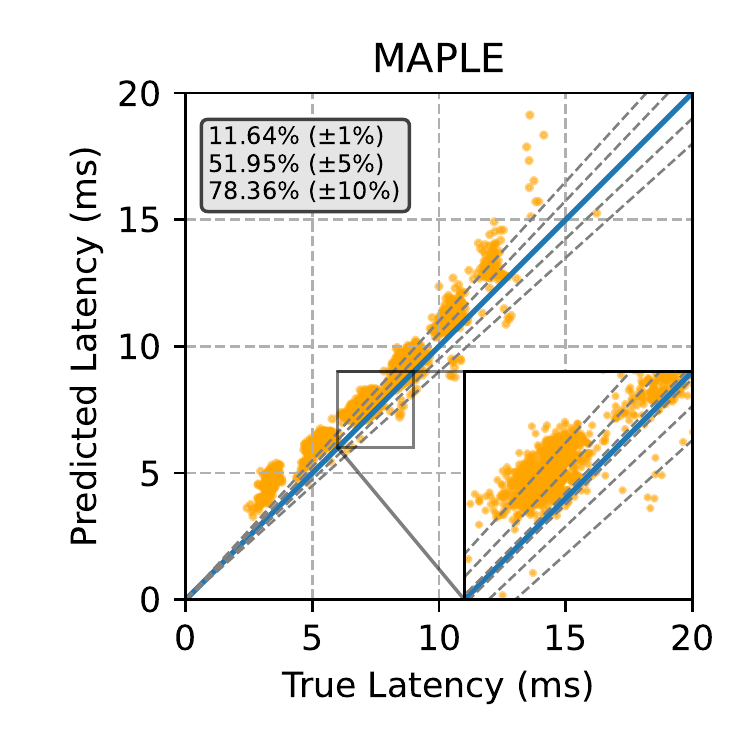}
\end{tabularx}
\caption{Visual comparison of true and predicted latency on Intel Xeon Gold 6230 and Nvdia RTX6000. The HELP and {\our} methods were trained on 900 samples collected from each of the seven devices in the training pool. The {\our} method was adapted by mixing in 3 samples whereas the HELP method was adapted using 10 samples. The blue line represents perfect prediction accuracy whereas the dashed gray lines represent $\pm1\%$, $\pm5\%$ and $\pm10\%$ error bounds. The error-bound accuracy for each technique is annotated in the top left corner. The insets provide a closer comparison of where each technique places the predicted latency. For example, for the Xeon 6230, HELP is able to place a significant number of points within $\pm5$ error but exhibits a considerable bias (many points are above the blue line). In contrast, {\our} distributes the predictions within $\pm5\%$ more evenly and is thus able to deliver an accuracy of 88.18\%.}
\label{fig:acc_visualize}
\end{figure}

While hardware performance counters have been widely available on various microprocessor architectures~\cite{cortexa9,amd64,intel} for over decade, they are scarcely found on GPUs or ASIC/FPGA based deep learning accelerators (DLA). Only some of the latest GPU models have begun to feature hardware performance counters. Moreover, since GPUs and DLAs feature significantly different hardware architectures, they tend to expose a different set of hardware counters ultimately leading to a different hardware descriptor. A model trained on CPU performance counters would have difficulty in interpreting a GPU-based hardware descriptor and would likely require a latency estimation model dedicated for GPUs or DLAs. Thus relying on device-specific hardware counters can potentially limit the versatility of the technique. Keeping the scarcity and versatility in mind, we propose to use CPU-specific counters to characterize GPUs. GPU performance strongly depends on how fast a microprocessor can continuously keep it fed with data, resulting in tight I/O coupling between the two devices. We propose to take advantage of this tight I/O coupling to characterize the GPU while measuring CPU performance counters, effectively using the CPU as a proxy to describe GPU behavior. This approach yields a versatile hardware descriptor that can be acquired on a wide variety of systems.

To assess if the proposed CPU-based descriptor is able to distinguish between different GPUs, we collect hardware descriptors from all devices used in this study and subsequently compute their euclidean distances. We use euclidean distance over cosine distances to ensure the magnitudes of the vectors are also take into account. We observe from Figure~\ref{fig:cosine} that all devices exhibit large distances from each other, suggesting that the hardware descriptors are sufficiently dissimilar. A key distance to note is between the Nvidia Titan X and TitanXP, which were paired with the same CPU (in the same PC). We note that despite using the same proxy device, the hardware descriptor describes the two GPUs differently (hence the large distance between the two GPUs). This discernability of the hardware descriptor is due to the tight I/O coupling between the CPU and GPU, which leads the CPU to behave differently with each GPU. Thus, employing CPU-specific hardware counters to yield a hardware descriptor results in a versatile technique that can target a wide variety of deep learning hardware.

\subsection{Regression Model Architecture}
The proposed method employs a compact neural network-based non-linear regression model for latency inference. The regression model is hardware-aware and DNN architecture-aware as it accepts the hardware descriptor and DNN architecture encoding as inputs. To cater for the discrete nature of the architecture encoding $\textbf{a}$ as well the continuous-valued hardware descriptor $\textbf{S}$, the regression model is designed as a dual-stream neural network architecture (Figure~\ref{fig:overview} provides an illustration). The first stream takes the architectural encoding as an input and is processed by two hidden layers. The architecture encoding is subsequently mapped to a 32-dimensional continuous space vector before being concatenated with the hardware descriptor $\textbf{S}$. The concatenated vector is the second stream of the regression model consisting of a further two hidden layers. Employing the dual-stream model enables the regression model to learn the characteristics of the architecture encoding and the hardware descriptor independently.
\section{Evaluation}
\subsection{Experiment setup}
To assess the efficacy of {\our}, we measure the accuracy of our predictor as well as HELP~\cite{lee2021help} and look-up table (LUT)-based approaches. Moreover, we employ an Oracle NAS approach to evaluate the identification of Pareto-optimal architectures.

\subsection{Comparison Metric}
Inspired by BRP-NAS~\cite{dudziak2020brp}, we employ error-bound accuracy as our primary metric, which is defined as the percentage of samples that falls within a given error-bound. In this study, we use $\pm 1\%$,  $\pm 5\%$,  and $\pm 10\%$ error-bounded accuracy. These metrics are widely used in the literature due to their interpretability. 

\subsection{Comparison baselines}
We compare the proposed {\our} against two well-known approaches i) LUT \cite{dai2019chamnet,cai2019once,wu2019fbnet,cai2018proxylessnas} and ii) HELP \cite{lee2021help}. Using a LUT for architecture latency yields a simple yet effective approach for estimating how a given architecture would perform on a new device. The number of measurements required depends only on the search space size and not on the number of architectures. Although our experiments showed that the accuracy of a LUT-derived latency depends highly on the complexity of the underlying hardware (see Figure~\ref{fig:acc_visualize}), we still consider an architecture latency LUT to be a strong baseline due to its simplicity, relative accuracy and convenience.

HELP~\cite{lee2021help} is a state-of-the-art method that can generalize to new hardware with as few as 10 samples. The authors demonstrated their technique with seventeen devices and collected 900 training samples. To eliminate any distribution bias, we train HELP and {\our} on the exact same training samples. Using the same samples ensures that the accuracy results are due to the technique's ability to generalize to new devices and not due to randomly being trained on a more representative distribution. We modify the HELP implementation to output error-bound accuracy instead of Spearman correlation.

\subsection{Efficacy of Few shot Adaptation to Unseen Hardware}
\label{sec:few-shot}
To compare the efficacy of the proposed approach with the competing methods, we begin by forming a training pool of seven devices (mentioned in section~\ref{sec:pipeline}) and using the eighth device for testing purposes. Importantly, we rotate devices into and out of the training pool using leave-one-out cross-validation and average the results. Using leave-one-out cross-validation ensures that the results are invariant of a specific device combination. We train both HELP and {\our} using the same 900 training samples per device and validate using all 15,625 architectures in NAS-Bench201. Table~\ref{tab:detail} shows a detailed comparison between HELP, the LUT baseline and our method. To stay consistent with the original study, we use 10 samples to adapt HELP to test devices \cite{lee2021help}. To demonstrate the hardware adaptation capability of {\our}, we evaluate our method with as few as 3 samples as well.

\begin{table*}
\scriptsize
\centering
\begin{tabularx}{1.0\textwidth}{lrrrrrrrrrr} 
\toprule
      &  No. of & \multicolumn{3}{c}{Unseen CPU}  & \multicolumn{5}{c}{Unseen GPU} & \multicolumn{1}{l}{}  \\
Method & \ Samples & i5-7600k & i9-9920k & Xeon 6230 & GTX1070 & RTX2080 & RTX6000 & TitanX & TitanXP   & Mean                  \\ 
\midrule
HELP   & 10 & 0.95 & 0.87 & 0.87 & 0.80 & 0.73 & 0.45 & 0.86 & 0.92 & 0.81 \\
{\our} & 3 & 0.92 & 0.90 & 0.88 & 0.95 & 0.79 & 0.83 & 0.84 & 0.84 & 0.87 \\
{\our}   & 10 & 0.97 & 0.92 & 0.92 & 0.97 & 0.90 & 0.86 & 0.95 & 0.94 & 0.93 \\
\bottomrule

\end{tabularx}
\caption{Comparison of few-shot adaptation efficacy between HELP and {\our}. Both techniques were trained with 900 points from each of the seven hardware devices in the training device-pool. HELP was adapted to the unseen devices by collecting 10 additional samples as suggested by the authors~\cite{lee2021help}. The efficacy of {\our} is demonstrated by mixing 3 as well as 10 samples. The reported metric is $\pm10\%$ error-bound accuracy. We note that although {\our} benefits from mixing in 10 samples, mixing only 3 samples also outperforms HELP algorithm and provides a 3\% improvement over HELP.}
\label{tab:detail}
\end{table*}

\begin{figure*}
\begin{tabularx}{\linewidth}{ccc}
\setlength{\tabcolsep}{0.01cm} 
     \includegraphics[width=0.47\linewidth]{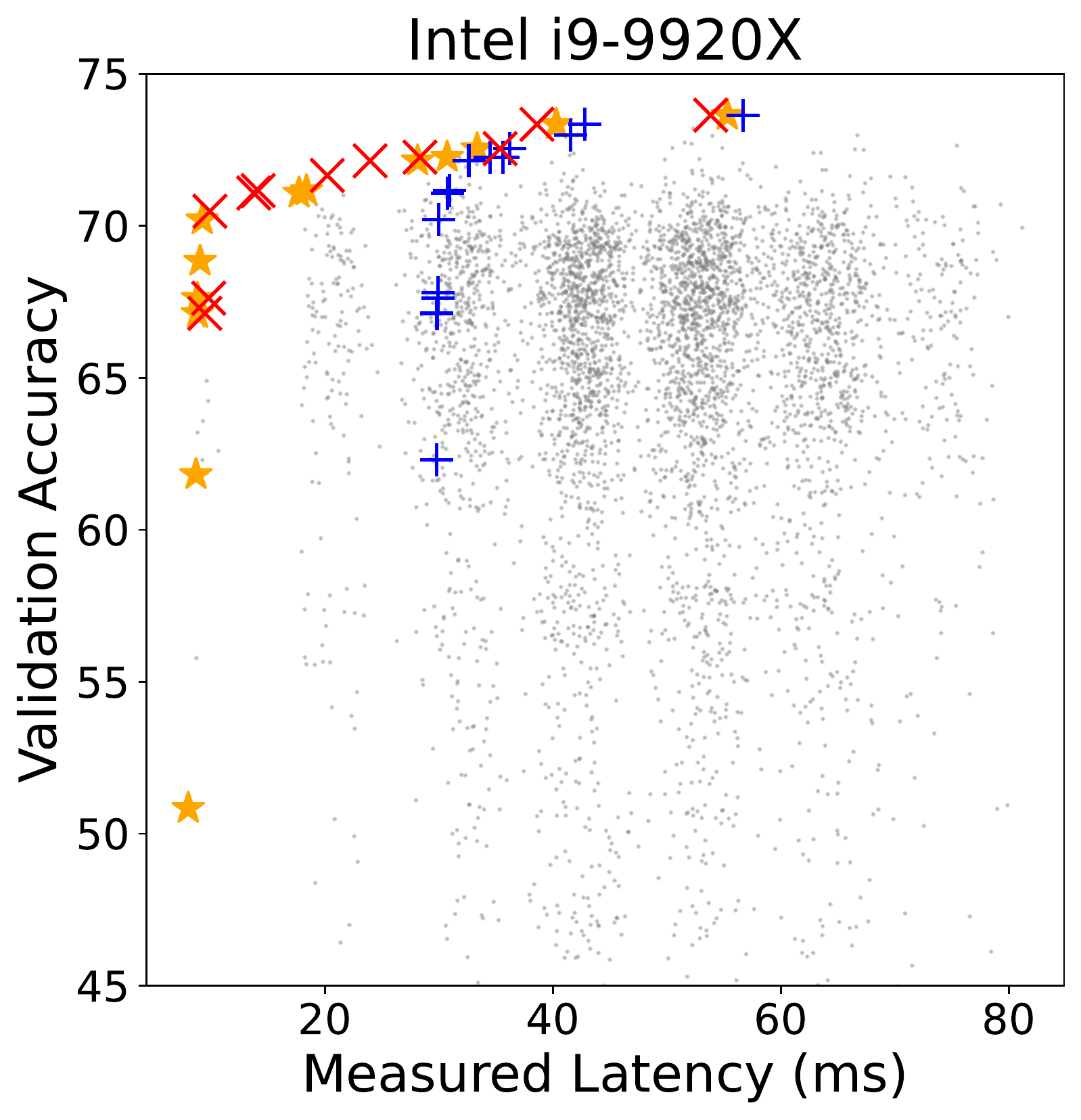} &  
     \includegraphics[width=0.47\linewidth]{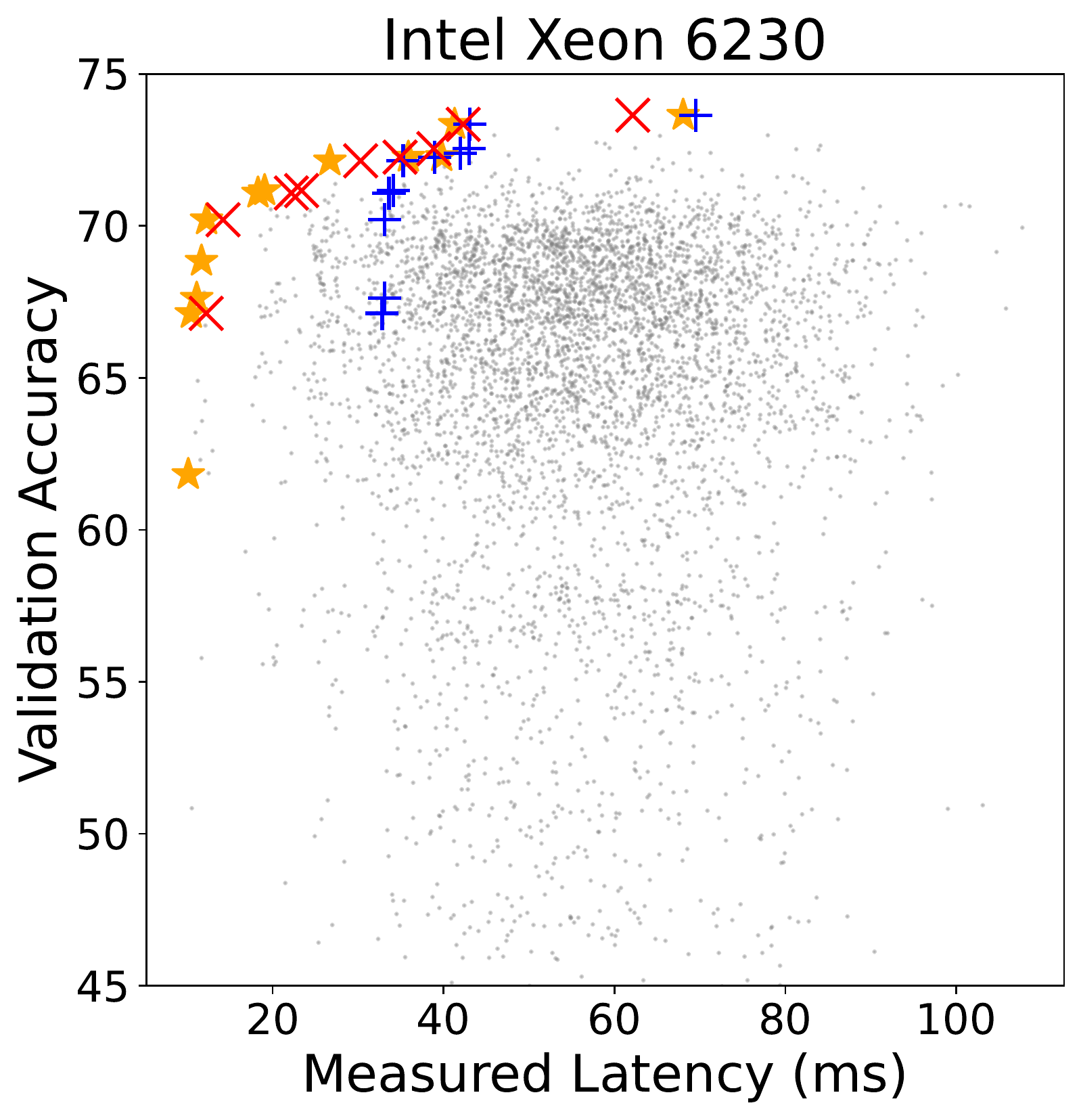} \\
     \includegraphics[width=0.47\linewidth]{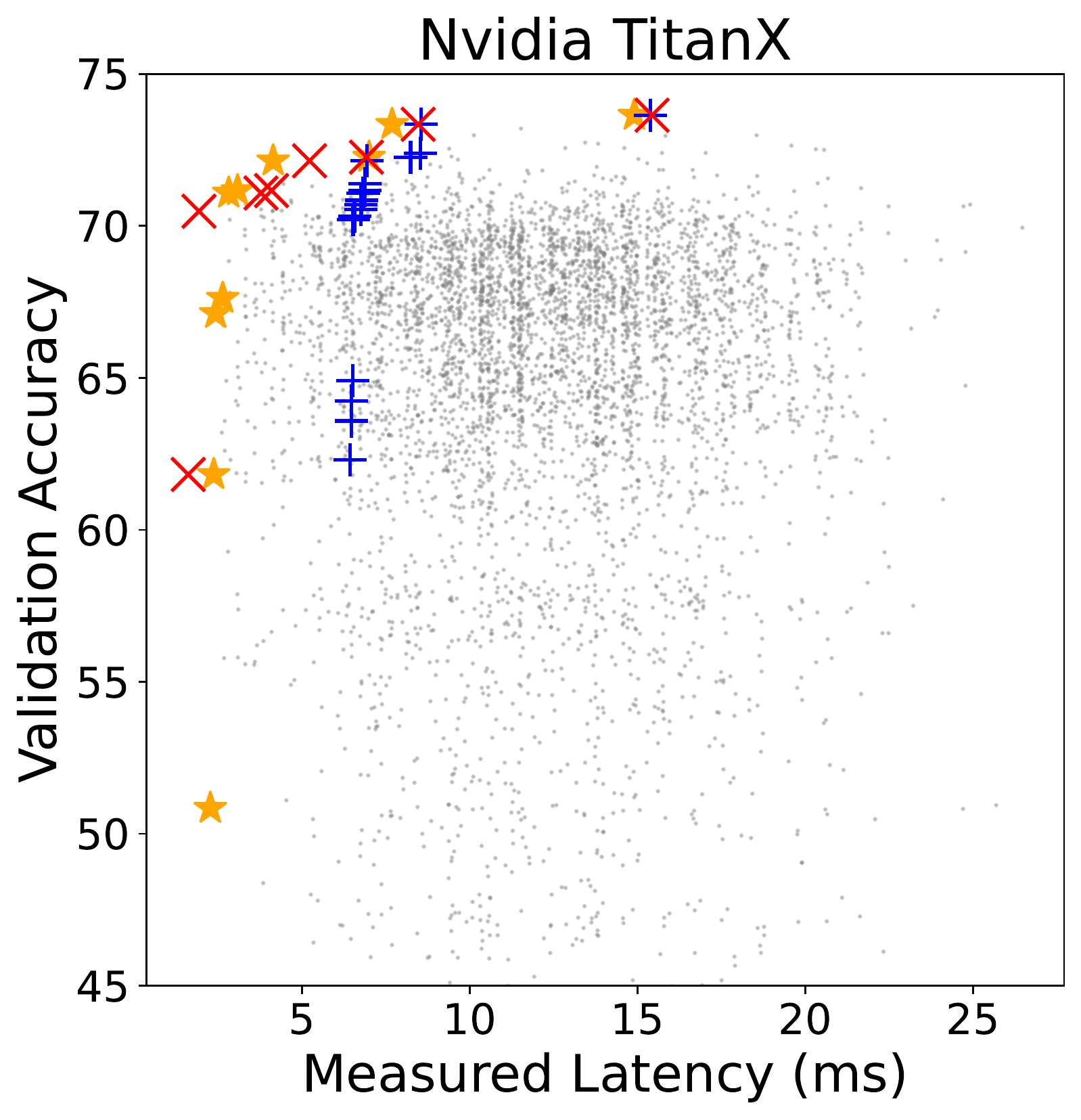} &
     \includegraphics[width=0.47\linewidth]{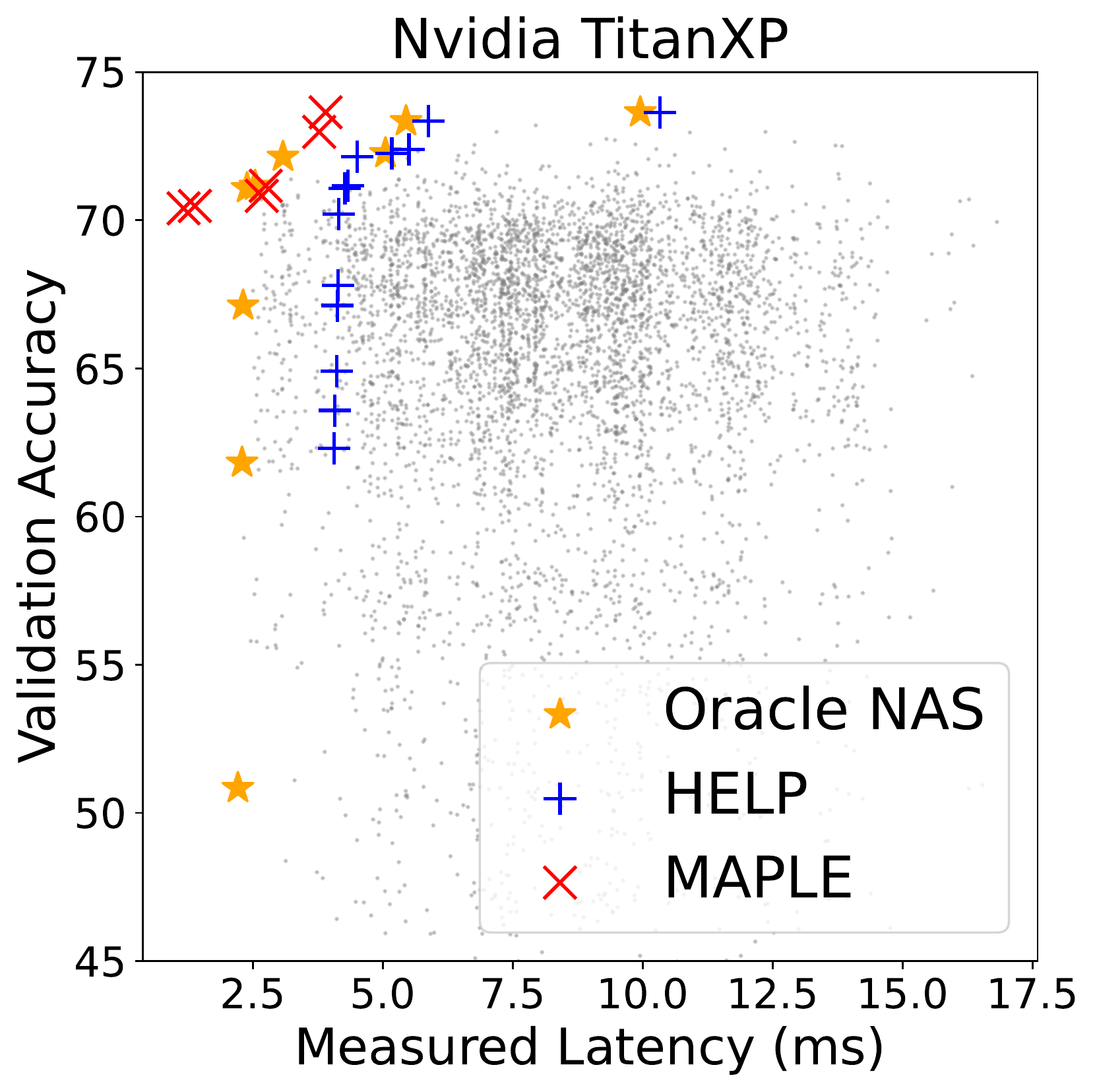}
     \end{tabularx}
\caption{Identification of Pareto-optimal architectures in NAS-Bench-201. Each point is a DNN architecture with CIFAR100 validation accuracy along with true latency (yellow star), MAPLE predicted latency (red cross) and HELP predicted latency (blue plus). We note that MAPLE only uses three adaptation samples whereas HELP uses the recommended ten. We observe that despite the lower number of adaptation samples, MAPLE identifies Pareto-optimal architectures more successfully than HELP.}
\label{fig:oracle}
\end{figure*}

Table~\ref{tab:detail} compares the $10\%$ error-bound accuracy between HELP and {\our}. This Table illustrates {\our}'s efficacy in rapid model adaptation. We note that {\our} reports an improvement of 6\% despite using only 3 samples for model adaptation compared to HELP's 10 samples. Moreover, we note that when {\our} uses 10 adaptation samples, the performance improves significantly from an average of 0.85 to 0.94. Importantly, our regression model was adapted to GPUs presented in Table~\ref{tab:detail} using CPU-based performance counters. The model is able to adapt to GPUs with only 3 samples due to the tight-coupling present between the CPU and GPU, as discussed in Section~\ref{sec:descriptor}. This illustrates the effectiveness of using widely available CPU performance counters to characterize GPU hardware.

Figure~\ref{fig:acc_visualize} provides an illustrative comparison between the architecture layer-wise predictor, HELP and {\our}. The blue line in the Figure serves as a guide for perfect prediction latency while the dashed gray lines visually show the $\pm 1\%$, $\pm 5\%$, and $\pm 10\%$ error bounds. We again use 10 samples with HELP and 3 samples with {\our} for model adaptation. In Figure~\ref{fig:acc_visualize} (top left) we show the correlation between the true latency and summed layer-wise latency. The layer-wise latency was computed by summing the execution time of all primitive operations that a given architecture was comprised of. Each sample was measured 25 times and averaged to get a robust estimate. We note that the layer-wise predictor consistently underestimates the architecture latency with the error growing for higher latency models. This illustrates that simply summing the operator latency fails to capture the intricacies of running a full DNN architecture. Employing HELP (top center) reduces the error significantly, with most architectures falling within $\pm 10\%$ of the actual latency. In contrast, {\our} is able to outperform HELP and exhibits an improvement of nearly 10\%, while utilizing only three samples for model adaptation. We observe a similar trend when assessing the plots for Nvidia RTX6000 GPU. The layer-wise predictor consistently overestimates the architecture latency. HELP again closes the gap between the predicted latency but generally underestimates the architecture latency for a significant number of samples. In comparison, {\our} can bring over half the points within $\pm 5\%$ error while employing just three points.

Finally, to visualize the identification of Pareto-optimal models we plot the true accuracy of each DNN architecture against its latency (Figure~\ref{fig:oracle}). We combine the true model accuracy with its true (measured) latency to effectively yield an Oracle NAS. The Oracle NAS represents the overall ground truth for every target device and identifies the true Pareto-optimal architectures. These true Pareto-optimal architectures are denoted via yellow stars in the Figure. Additionally, we combine the true model accuracy with predicted latency to identify DNN architectures that {\our} and HELP would delineate as Pareto-optimal. Several important observations can be made from Figure~\ref{fig:oracle}. First, the true Pareto-optimal architectures are different for each target device, strongly suggesting device dependency for DNN latency. Secondly, despite using just three adaptation points, {\our} identifies DNN architectures much closer to the true Pareto-optimal models than HELP which requires ten adaptation samples. Finally, we note that even if the prediction accuracy is relatively close (e.g. for Titan X), it does not necessarily guarantee successful identification of optimal models in the search space.

\section{Conclusions}
In this work, we proposed {\our}, a simple yet effective latency predictor that is able to rapidly adapt to new hardware. {\our} is based on a novel device descriptor that is able to characterize the target hardware by measuring ten key performance metrics, including cache efficacy, computational rate and instruction count, among others. These metrics are captured by measuring key CPU-based hardware performance counters while executing key primitive workloads on top. Hardware performance counters yield an efficient hardware descriptor that enables {\our} to generalize to new devices with as few as three samples. Moreover, the proposed hardware descriptor is also able to characterize GPUs despite being based on CPU-based hardware performance counters. The GPU characterization is possible as the proposed technique takes advantage of the tight-coupling present between the CPU and GPU. In contrast to other approaches which use fine-tuning or meta-learning, we incorporated the target device characterization at training time, yielding a simple yet accurate approach to latency prediction. We validated {\our} by conducting a series of experiments. First, we assessed the latency prediction accuracy on new hardware with as few as three and ten samples. Employing just three adaptation samples yielded a 6\% improvement over the state of the art while using ten samples yielded an improvement of 12\%. Second, we assessed how many samples  the proposed method requires to generalize effectively to new devices. We found that {\our} yielded an average improvement of 8-10\% over other approaches. Finally, compared to the state-of-the-art techniques, {\our} also requires significantly fewer training examples to generalize to unseen network architectures. These characteristics yield a simple latency predictor that can significantly reduce the cost hardware-aware NAS.

\medskip

{
\small
\printbibliography
}

\end{document}